\documentclass[letterpaper, 10 pt, conference]{ieeeconf}  

\IEEEoverridecommandlockouts                     
\usepackage{balance}
\usepackage{graphicx}       
\usepackage{amsmath}        
\usepackage{amssymb}      
\usepackage{mathptmx}       
\usepackage{times}          
\usepackage{booktabs}      
\usepackage{pifont}         
\usepackage{tabularx}   
\usepackage{threeparttable}
\usepackage[caption=false,font=footnotesize]{subfig}
\newcolumntype{Y}{>{\raggedright\arraybackslash}X}

\usepackage[table]{xcolor} 
\usepackage{cite}    
\usepackage[hidelinks]{hyperref}
\usepackage{multirow}
\title{\LARGE \bf
TEGA: A Tactile-Enhanced Grasping Assistant for Assistive Robotics via Sensor Fusion and Closed-Loop Haptic Feedback}

\author{Hengxu You$^{1*}$, Tianyu Zhou$^{1*}$, Fang Xu$^{1}$, Kaleb Smith$^{2}$ and Eric Jing Du$^{1}$%
\thanks{*This work was supported by the NVAITC and UFIT, University of Florida.}
\thanks{*These authors contributed equally.}
\thanks{$^{1}$All authors are with the Department of Civil and Coastal Engineering, University of Florida, Gainesville, FL 32611, USA.
        {\tt\small \{you.h, zhoutianyu, xufang\}@ufl.edu}; {\tt\small eric.du@essie.ufl.edu}}%
\thanks{$^{2}$All authors are with Nvidia }%
\thanks{Corresponding author: Eric Jing Du}%
}
\begin{document}

\maketitle

\thispagestyle{empty}
\pagestyle{empty}

\begin{abstract}

\textbf{Recent advances in teleoperation have enabled sophisticated manipulation of dexterous robotic hands, with most systems concentrating on guiding finger positions to achieve desired grasp configurations. However, while accurate finger positioning is essential, it often overlooks the equally critical task of grasp force modulation—vital for handling objects of diverse hardness, texture, and shape. This limitation poses a significant challenge for users, especially individuals with upper-limb disabilities who lack natural tactile feedback and rely on indirect cues to infer appropriate force levels. To address this gap, We present the tactile-enhanced grasping assistant (TEGA), a closed‑loop assistive teleoperation framework that fuses EMG‑based intent‑to‑force inference with visuotactile sensing mapped into real‑time vibrotactile feedback via a wearable haptic vest, enabling intuitive, proportional force adjustment during manipulation. A wearable haptic vest delivers real-time tactile feedback, allowing users to dynamically refine grasp force during manipulation. User studies confirm that the system substantially improves grasp stability and task success, underscoring its potential for assistive robotic applications.}

\end{abstract}

\section{INTRODUCTION}

Assistive robotics has advanced robotic manipulation significantly in recent years, offering innovative solutions for individuals with upper limb disabilities \cite{bauer2020review}. Modern assistive systems empower users to control robotic hands that augment or even replace lost hand functionality, enabling them to perform complex manipulation tasks that are otherwise challenging or impossible \cite{bittencourt2024initial}. For many users, the ability to execute reliable and dexterous grasping tasks is transformative; such technology increases independence, reduces reliance on caregivers, and dramatically improves overall quality of life \cite{hansen2024multimodal}.

Despite notable advances, most assistive robotic systems continue to emphasize accurate finger positioning, while under-addressing the requirement for dynamic and adaptive force control, which is a recognized limitation in the latest tactile manipulation and policy learning research \cite{hogan2020tactile}. Although precise positioning is crucial, another equally important challenge remains largely unaddressed: the dynamic adjustment of grasp force. When interacting with objects of varying hardness and weight, fine-tuning the force applied during grasping is essential to prevent slippage or damage \cite{zhang2022design}. This challenge is especially pronounced for users with limited or absent hand function, who lack natural tactile feedback. Without a mechanism to perceive and adjust force in real time, even the most advanced positioning systems fall short of providing truly dexterous control \cite{aug2022haptic}.

Recent state-of-the-art designs have incorporated multimodal sensory tracking (eye tracking and motion tracking) to better infer user intention and create a smoother, more adaptable control interface \cite{wang2024visual}. These enhancements have reduced the cognitive load on users and improved control fidelity; however, they still do not supply the critical sensory feedback needed to modulate grasp force during object manipulation.

\begin{figure*}[ht!]
    \centering
    \includegraphics[width=\textwidth]{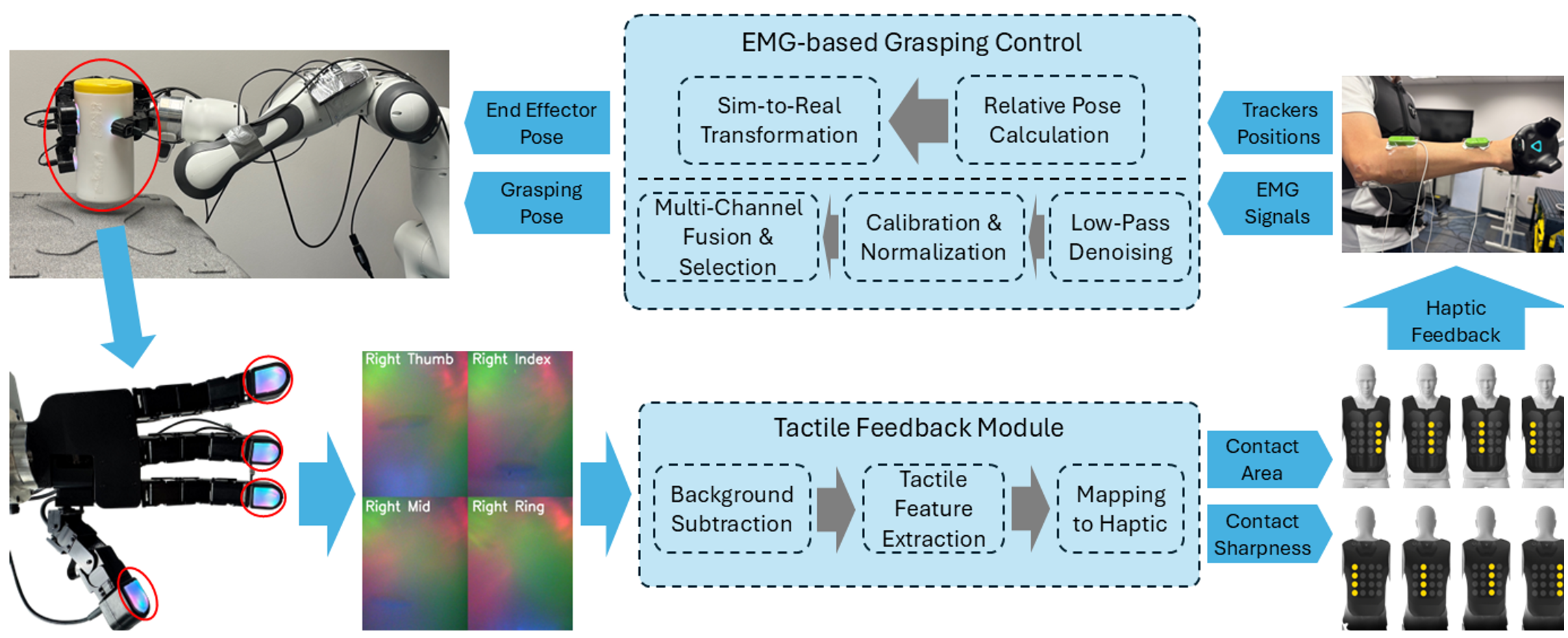} 
    \caption{
    \textbf{Overview of the Tactile-Enhanced Grasping Assistant (TEGA) system.}  The system integrates EMG-based force inference, visuo-tactile sensing, and haptic feedback to enable intuitive, precise, and adaptive grasping force control through real-time closed-loop interaction. EMG signals from the user’s upper arm muscles are processed to infer grasp intent, while tactile sensor data from the robotic fingertips is converted into real-time vibrotactile feedback delivered via a wearable haptic vest.
    }
    \label{fig:1}
\end{figure*}

To enable dexterous control, a closed-loop system is required that not only commands precise grasping motions but also delivers real-time tactile sensory feedback \cite{ding2021sim}. Such feedback is vital for simulating the natural sensations of a complete hand, enabling users to "feel" contact, pressure, texture, and slip. 

In response to these challenges, we propose the tactile-enhanced grasping assistant (TEGA), an assistive robotic system designed to empower users with limited or absent hand function. Fig.~\ref{fig:1} demonstrates the overall workflow of the proposed system. TEGA features a dual-loop architecture that integrates multiple sensing and control modalities to bridge the gap between user intent and robotic execution. An adaptive model interprets electromyography (EMG) signals from the upper arm muscles to infer the user’s intended grasp force \cite{uthayakumar2020teleoperation}. In parallel, a complementary projection model processes tactile sensor data from the robotic hand and converts it into distinct vibration patterns \cite{steinbach2018haptic}. These vibration patterns are delivered to the user via the haptic vest. This integrated, closed-loop approach continuously guides the robot’s grasping motion and adjusts the applied force based on real-time sensory input, thereby enhancing grasp stability and overall safety.
The main contributions of this work are as follows:
\begin{itemize}

    \item  We present \textbf{a novel assistive robotic system} that integrates EMG-based force inference with real-time tactile feedback, enabling users to intuitively modulate grasp force during complex manipulation tasks.
    \item  We develop \textbf{a multi-modal sensing framework} that combines augmented reality-based pose tracking with visuo-tactile sensing, capturing rich user and environmental data to enhance control precision, adaptability, and context-awareness.
    \item We propose a set of \textbf{projection models} that translate both \textbf{EMG signals} and \textbf{tactile sensor} data into actionable control commands and haptic feedback cues, thereby bridging the gap between user intent and robotic actuation in a closed-loop system.
    
\end{itemize}


\section{Background and Related Work}
\subsection{Tactile Feedback for Assistive Grasping}
Recent advances in tactile sensing have enhanced robotic manipulation, especially in assistive grasping. Vision-based tactile sensors can capture contact properties, force distribution, and surface textures, enabling adaptive systems that dynamically adjust grip strength from feedback.\cite{lambeta2020digit,padmanabha2020omnitact}.Beyond autonomous grasping, tactile sensing is vital in human-in-the-loop control. Touch enables dexterous manipulation through texture, weight, and slip cues~\cite{cremer2020skinsim}. Haptic gloves provide operators with real-time sensations of force and texture~\cite{caeiro2021systematic}, aiding users with reduced proprioception. By mapping robotic tactile data to glove actuators, users can remotely ``feel'' contact forces and surface properties~\cite{jafari2016haptics}.

Building on these sensing and feedback advances, recent robotic policy learning methods have shown that combining visual and tactile data with diffusion policies enables robust adaptation and recovery from manipulation failures \cite{chi2023diffusion}. Slow-fast architectures further improve responsiveness by integrating low-latency tactile reflexes \cite{xue2025reactive}. Together, these insights inform TEGA’s dual-loop design, which unifies operator-in-the-loop learning with automated force control, pointing to future directions in adaptive and resilient manipulation.

While haptic feedback technologies have advanced significantly, enabling human operators to feel robotic interactions more intuitively, individuals with sensory impairments or upper limb disabilities face unique challenges in utilizing such systems effectively. Many commercial haptic devices assume the user has full tactile sensation, meaning users with reduced or absent tactile perception may not benefit from force feedback or vibration-based cues in the same way as an able-bodied operator \cite{abd2022multichannel}. 

\subsection{EMG-Based Grasp Assistance}
EMG has emerged as a promising interface for robotic control, particularly in assistive applications where users may have limited mobility but retain residual muscle activity. EMG-based systems enable intent-driven grasp control by interpreting muscle signals and translating them into corresponding robotic actions \cite{tacca2024wearable,sierotowicz2022emg}. Recent advancements in EMG signal processing have further improved the stability and responsiveness of these systems. Traditional methods relied on threshold-based activation, where users contract specific muscles to trigger discrete actions. However, more recent approaches leverage adaptive filtering and multi-channel EMG classification to enable continuous and proportional control of robotic limbs, allowing users to modulate grasp force dynamically \cite{xu2022continuous}.

Despite its advantages, EMG-based control alone lacks the instantaneous tactile feedback inherent to biological grasping, where humans naturally perceive grip force, slippage, and contact stability in real time. Without appropriate haptic feedback, users must rely solely on visual observation, leading to delays and inaccuracies in grasp force modulation and making fine motor adjustments more challenging \cite{bozzacchi2016grasping}. This is particularly important for individuals with hand disabilities, as visual observation alone may not provide sufficient real-time information about grip stability, contact force, or surface texture. Without an alternative means of perceiving these critical grasping cues, users must rely on trial and error, which can lead to instability in manipulation and reduced efficiency in performing tasks \cite{jafari2016haptics}. 

Thus, developing an alternative method that closely mimics real hand sensation could significantly improve operational quality for users with hand disabilities. By replicating the nuanced sensory experiences of natural touch, users can achieve more precise and stable grasping movements, bridging the gap between biological and artificial tactile perception. These advancements reinforce the role of haptic feedback as a critical tool in assistive robotics, ensuring that individuals with hand loss or sensory deficits can interact with their environment naturally and regain functional independence.

\begin{figure}[t]
    \centering
    \includegraphics[width=\columnwidth]{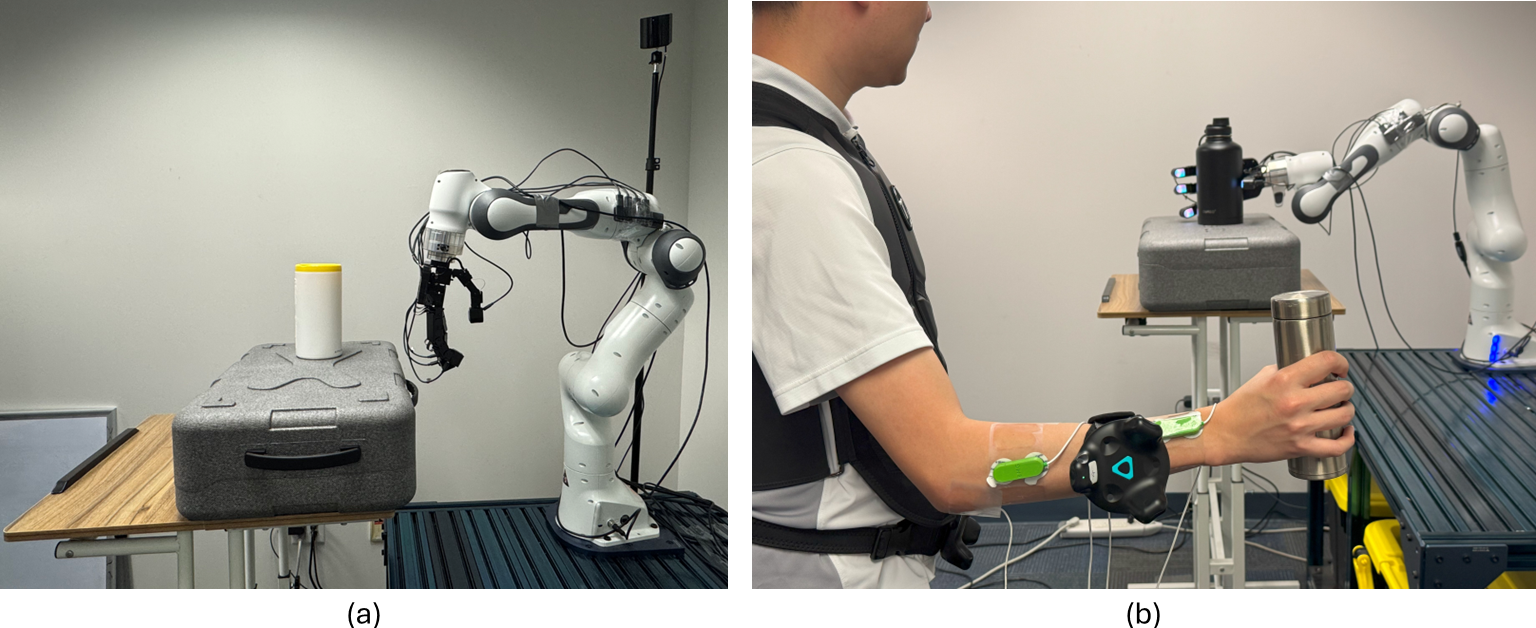}
    \caption{\textbf{Hardware devices setup.} 
        (a) robot arm-hand system with mounted DIGIT sensors and (b) human operator with haptic vest, motion trackers and EMG sensors.}
    \label{fig:2}
\end{figure}

\section{SYSTEM DESIGN}
Our proposed system adopts a multimodal assistive robotic architecture to support dexterous grasping for users with upper-limb disabilities (Fig.~\ref{fig:2}). The platform consists of an Allegro Hand mounted on a Franka Emika Panda robot arm, with each finger equipped with a DIGIT tactile sensor. A HoloLens headset tracks upper-limb endpoint poses for robot control, while three EMG electrodes placed on distinct muscle groups (Sensors 1–3) capture muscle activity to infer user grasp intent. A bHaptics vest (TactSuit) provides real-time feedback derived from tactile sensing. Together, these components form a closed-loop system integrating visual, tactile, and bioelectrical signals to improve grasp stability and adaptability.

The vibrotactile vest serves as a sensory substitution channel that conveys task-relevant contact state information. Tactile features extracted from the DIGIT sensors, including relative force intensity and deformation trends, are mapped to spatially distributed vibration cues on the torso. While users with reduced or absent hand sensation may lack fine fingertip perception, vibration perception over larger body areas often remains available, enabling closed-loop adjustment of grasping force based on these cues.

\begin{figure}[t]
    \centering
    \includegraphics[width=\columnwidth]{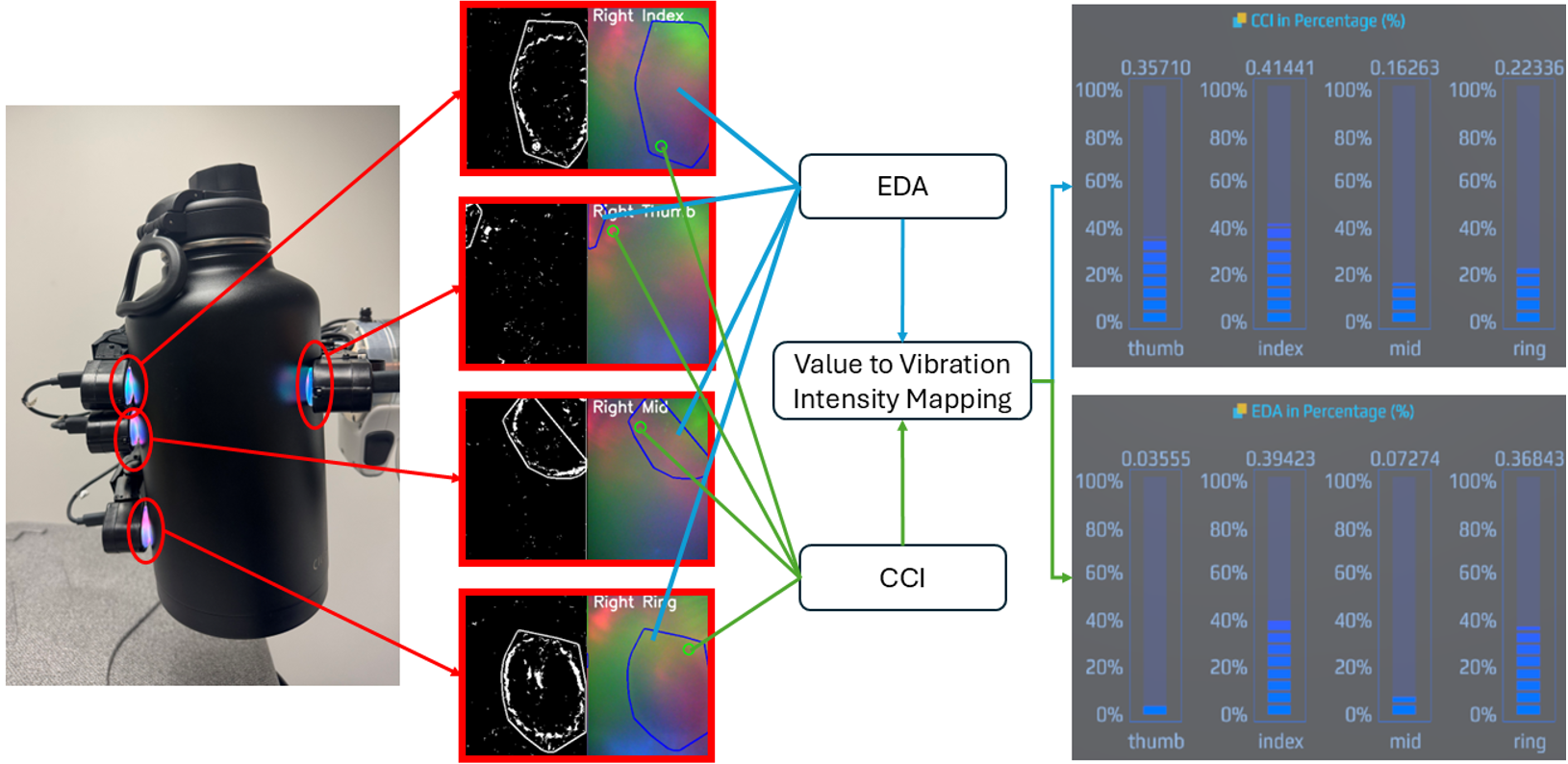}
    \caption{\textbf{Visualization of CCI and EDA.} 
        The grayscale image is extracted from the raw RGB tactile image to estimate the depth map of the deformation. The green circle marks the pixel point with the highest Contact CCI, representing the location with the most concentrated pressure. The blue circle identifies the sharpest pressure point, encompassing EDA, which represents the spatial extent of the deformation. Both CCI and EDA values are normalized within the range [0,1] and transmitted to the haptic vest, where they modulate the intensity of the corresponding vibration unit.}
    \label{fig:3}
\end{figure}

\subsection{Tactile Feedback Module}
\label{sec:tactile_feedback_module}

Rather than attempting to replicate fine fingertip tactile sensations, the proposed vibrotactile vest is designed as a sensory substitution channel that conveys task-relevant contact state information. Specifically, tactile features extracted from the robotic fingertips, such as relative force intensity and deformation trends, are mapped to spatially distributed vibration cues on the torso. While users with reduced or absent hand sensation may lack fine tactile perception at the fingertips, vibration perception over larger body areas often remains available. This mapping enables users to perceive changes in contact state and adjust grasping force in a closed-loop manner, even without direct fingertip sensation.

For real-world human touch tactile feedback, a critical perspective must be considered: Contact Morphology Metrics. \cite{bhirangi2409anyskin, son2024behavioral} explore the use of tactile sensors to capture dynamic contact information, emphasizing the role of stress concentration detection and dynamic force transfer in understanding contact interactions. Inspired by these findings, we introduce Contact Morphology Metrics (CMM) as a refined approach to systematically quantify local contact geometry between a fingertip and an object. Specifically, we define two complementary metrics: Contact Concentration Index (CCI), which captures localized pressure distribution to assess stress concentration patterns, and Effective Deformation Area (EDA), which measures the extent of contact spread to characterize force dispersion over the contact surface. The CCI quantifies the depth of deformation by measuring how sharply the pressure is concentrated around the maximum pressure point, which indicates a deep, localized indentation typically produced by a sharp edge or point contact. EDA quantifies the spatial extent of the deformation by counting the number of pixels in the pseudo-pressure map that fall within the upper 90\% of the pressure values. A larger EDA reflects a broad, distributed contact, while a smaller EDA signifies a more localized contact \cite{pinto2024hardware}. 

Fig.~\ref{fig:3} illustrates the effect of EDA and CCI when grasping different objects. Together, CCI and EDA offer complementary insights into the tactile interaction CCI captures the intensity (or depth) of contact, and EDA reveals its spatial spread. For example, a high CCI with low EDA signifies a focal contact that may require only a slight force adjustment to prevent damage or instability, while a lower CCI combined with higher EDA indicates a broader contact that can tolerate larger force increments without risking slippage.

Rather than replicating fine fingertip tactile sensations, the vibrotactile vest is designed as a sensory substitution channel that conveys task-relevant contact state information. Tactile features extracted from the robotic fingertips, such as relative force intensity and deformation trends, are mapped to spatially distributed vibration cues on the torso. While users with reduced or absent hand sensation may lack fine fingertip perception, vibration perception over larger body areas often remains available, enabling closed-loop adjustment of grasping force based on these cues.

To quantify tactile interactions, we adopt Contact Morphology Metrics (CMM) inspired by prior work on stress concentration and dynamic force transfer \cite{bhirangi2409anyskin, son2024behavioral}. Specifically, we define two complementary metrics: Contact Concentration Index (CCI), which captures localized pressure concentration to characterize deformation depth, and Effective Deformation Area (EDA), which measures contact spread to describe force dispersion over the surface. CCI reflects sharp, localized indentation, while EDA is computed as the number of pixels within the upper 90\% of pseudo-pressure values, indicating the spatial extent of contact \cite{pinto2024hardware}.

Fig.~\ref{fig:3} illustrates the effects of CCI and EDA across different objects. Together, these metrics provide complementary insights: CCI captures contact intensity, whereas EDA reflects spatial spread. For example, high CCI with low EDA indicates focal contact requiring careful force regulation, while lower CCI with higher EDA corresponds to broader contact that tolerates larger force variations.

\subsubsection{EDA and CCI Calculation}
\label{sec:eda_and_cci_calculation}
To change the default, adjust the template as follows. To isolate dynamic contact-induced deformation, we subtract a pre-recorded baseline tactile image from the real-time DIGIT stream. Let $I_i^t(x,y)=[r,g,b]^T$ denote the RGB tactile image of the $i$-th finger at time $t$, and $I_i^{base}(x,y)$ the corresponding baseline image. The differential image is defined as $I_i^{\prime t}=I_i^t-I_i^{base}$.  

The grayscale pressure proxy map is then computed as:
\begin{equation}
I_{i,gray}^t(x,y)=\omega_r r+\omega_g g+\omega_b b,
\end{equation}
where $\omega_r,\omega_g,\omega_b$ are fixed weighting coefficients satisfying $\omega_r+\omega_g+\omega_b=1$.

The resulting map $I_{i,g}^{t}(\Omega)$ could be treated as an estimation of the pressure map during interaction and the pressure weighted centroid across all pixels is calculated by:
\begin{equation}
\mu^{t}_{i,x} = \frac{\int x \cdot I_{i,gray}^t(x,y)\,dx\,dy}{\int I_{i,gray}^t(x,y)\,dx\,dy}, \quad
\mu^{t}_{i,y} = \frac{\int y \cdot I_{i,gray}^t(x,y)\,dx\,dy}{\int I_{i,gray}^t(x,y)\,dx\,dy}.
\label{eq:2}
\end{equation}

The standard deviation of the pressure distribution is calculated as:
\begin{equation}
\sigma_{i}^{t} =
\sqrt{\frac{\int \Big[(x - \mu_{i,x}^{t})^{2} + (y - \mu_{i,y}^{t})^{2}\Big] \cdot I_{i,\text{gray}}^{t}(x,y)\,dx\,dy}
{\int I_{i,\text{gray}}^{t}(x,y)\,dx\,dy}}.
\label{eq:3}
\end{equation}

According to (\ref{eq:2}) and (\ref{eq:3}), the threshold to select the effective deformation area could be determined by:
\begin{equation}
T_{i}^{t} = I_{i,\text{gray}}^{t}\!\left(\mu_{i,x}^{t}, \mu_{i,y}^{t}\right) - 1.28\,\sigma_{i}^{t}.
\label{eq:4}
\end{equation}

Then the EDA of current frame could be adaptive determined by:
\begin{equation}
EDA_{i}^{t} = \sum_{(x,y)\in I_{i,gray}^{t}} 
\mathbf{1}\!\left\{ I_{i,\text{gray}}^{t}(x,y) > T_{i}^{t} \right\}.
\label{eq:5}
\end{equation}

The coefficient 1.28 was empirically selected from preliminary experiments as a robust threshold for suppressing background noise while preserving dominant contact-induced deformation. Based on (\ref{eq:5}), the CCI for the current frame is calculated as:
\begin{equation}
CCI_{i}^{t} = \frac{\max_{(x,y)\in \Omega} I_{i,gray}^{t}(x,y)}{EDA_{i}^{t}}.
\label{eq:6}
\end{equation}

\subsubsection{Mapping to Haptic Vest}
\label{sec:mapping_to_haptic_vest}
The haptic feedback provided by the haptic vest is central to our system’s ability to deliver intuitive tactile cues. The TactSuit X40 \cite{simonsson2022configurable} is applied as the haptic devise which consists of 32 vibration units arranged in two 4×4 arrays with16 units on the front and 16 on the back. These are organized into four columns pairs $(c_1,c_2,c_3,c_4)$. Each corresponding pair, where the leftmost unit on the front is paired with the leftmost unit on the back and so on across the columns, is designated for a single finger with thumb, index, middle, and ring, respectively. The goal is to map the tactile measurements, EDA and CCI, to the vest’s vibration parameters of $c_1$ to $c_4$. The front 16 units vibration intensities are used to reflect the CCI, while the back 16 units vibration intensities reflect the EDA. 

We firstly calibrate each DIGIT sensor to determine the maximum observed values of EDA and CCI. For finger i with its corresponding $c_i$, we get the possible maximum value of EDA and CCI as $eda_i^{max}$ and $cci_i^{max}$. To ensure the normalization and stability of output, we apply the sigmoid function as the mapping function and use the calibrated values for adjustment. For CCI, the mapping function is:
\begin{equation}
\sigma_{i}^{CCI}(CCI_{i}^{t}) =
\frac{1}{1 + \exp\!\left(-k_{CCI}\cdot\left(CCI_{i}^{t} - \frac{cci_{i}^{\max}}{2}\right)\right)}.
\label{eq:7}
\end{equation}

To ensure that the linear region starts at approximately $CCI_i^t=0$ ($\sigma_i^{CCI}(0)\approx0.01$) and ends at $CCI_i^t=cci_i^{max}$ ($\sigma_i^{CCI}(cci_i^{max})\approx0.99$), the value of $k_{CCI}$ is selected as:
\begin{equation}
\sigma_{i}^{CCI}(0) =
\frac{1}{1 + \exp\!\left(-k_{CCI}\left(0 - \tfrac{cci_{i}^{\max}}{2}\right)\right)}
= 0.01.
\label{eq:8}
\end{equation}

Given that the intensity range of haptic suite is 0-100, we further scale up the output as $p^t_{i,front}=100\cdot\sigma_i^{CCI}(CCI_i^t)$. Similarly, EDA is mapped to intensity as:
\begin{equation}
\sigma_{i}^{EDA}(EDA_{i}^{t}) =
\frac{1}{1 + \exp\!\left(-k_{EDA}\cdot\left(EDA_{i}^{t} - \tfrac{eda_{i}^{\max}}{2}\right)\right)}.
\label{eq:9}
\end{equation}

\subsection{EMG-based Grasping Pose Estimation}
\subsubsection{Preprocess of raw EMG inputs}
To robustly estimate a user’s intended grasp force, the proposed system employs a rule-based mapping of three-channel EMG signals from the forearm. Using multiple channels ensures stability and reliability, which mitigate noise and variability in electrodes. The electrodes are placed on key muscle groups including Flexor Digitorum Profundus (FDP), Flexor Digitorum Superficialis (FDS), and Flexor Pollicis Longus (FPL) \cite{jones1997incidence}. 

To reduce noise in the high-frequency EMG signal, we preprocess the streaming data using a 4th-order Butterworth low-pass filter with a 50 Hz cutoff frequency, optimized for human-perceptible comfort. To further enhance stability, we average the filtered signal $\tilde{E}_{l}$ over every $\tau$ samples, retaining the resulting values as the downsampled signal. Specifically for each channel $i\in{FDP,FDS,FPL}$, the original sampling frequency is m Hz, and the new effective frequency is calculated as $f_i^d=\frac{m}{\tau}$. The downsampled values is then computed by averaging every $\tau$ samples as:

\begin{equation}
\bar{E}_{l}(t) = \frac{1}{\tau} 
\sum_{k=(t-1)\cdot\tau + 1}^{j \cdot \tau} 
\tilde{E}_{l}(k).
\label{eq:10}
\end{equation}
A calibration process is then conducted based on  $\bar{E}_{l}(t)$ to determine the maximum possible value for each channel, denoted as $\bar{E}_{l}^{max}$. The averaged signal $\bar{E}_{l}(t)$ is then discretized into five equal intervals with a range of $\frac{\bar{E}_i(t)}{4}$, corresponding to five pre-defined poses ${p_1,p_2,p_3,p_4,p_5}$, where a higher interval represents a tighter, more squeezed grasp. Let $\alpha_i(t)$ denotes the selected pose index for channel $i$ at time $t$, then the mapping rule from  $\bar{E}_{l}(t)$ to poses is:

\begin{equation}
\alpha_{i}(t) = \left\lfloor 
\frac{4 \cdot \bar{E}_{l}(t)}{\bar{E}_{i}^{\max}} + 1
\right\rfloor.
\label{eq:11}
\end{equation}

\begin{figure}[t]
    \centering
    \includegraphics[width=\columnwidth]{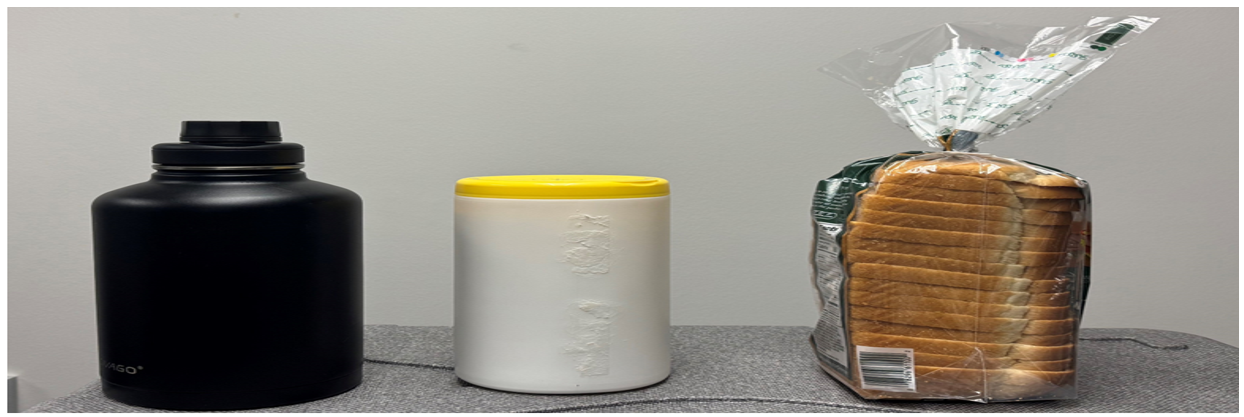}
    \caption{\textbf{Target objects to be tested.} Object 1 (water bottle, rigid and heavy), Object 2 (plastic wet wipes container, medium stiffness), and Object 3 (bag of bread, soft and deformable).}
    \label{fig:4}
\end{figure}

\subsubsection{Fusion of channel outputs}
With the selected pose indices from all three channels, the final grasping pose is determined by majority voting and average voting. Specifically, if at least two channels agree on the same index, then this index will be the final pose index as:
\begin{equation}
\bar{\alpha}(t) = c \;\; \text{if} \;\; \exists\, i,j \;\; \big( \alpha_{i}(t) = \alpha_{j}(t) = c, \; i\neq j \big).
\label{eq:12}
\end{equation}

If all three channels give different index values, then the final pose index is determined as:
\begin{equation}
\bar{\alpha}(t) =
\left\lfloor 
\frac{\alpha_{1}(t) + \alpha_{2}(t) + \alpha_{3}(t)}{3}
\right\rfloor.
\label{eq:12}
\end{equation}

\section{Experiment}

\subsection{Test System Setup}
This case study evaluates whether the proposed VR-EMG-haptic teleoperation system enables adaptive grasping force control, allowing users to effectively manipulate objects of varying material properties with precision and safety. To assess the system’s effectiveness, we examine its ability to support two critical aspects of grasp force control: For rigid objects, the system should provide sufficient gripping force to ensure a secure hold, preventing unintended dropping during object handling. For soft and deformable objects, the system should enable users to apply minimal yet stable grip force, preventing slippage while avoiding over-compression that could cause invisible deformation or structural damage.

The system was tested using a set of everyday objects with distinct material properties and deformability, including a water bottle (heavy and rigid, denoted object 1), a plastic wet wipes container (medium weight and softness, denoted object 2), and a bag of bread (light and soft, denoted object 3) (Fig.~\ref{fig:4}). The task involved a pick-and-place operation between two predefined start and end locations, with the gripping interaction recorded and analyzed based on three performance metrics: task completion time, grip success rate and slippage count. We labeled a trial as unsuccessful trial when the object dropped. For object 3, we also count the number of extreme deformation caused by unnecessary strong force. 

\begin{figure}[t]
    \centering
    \includegraphics[width=\columnwidth]{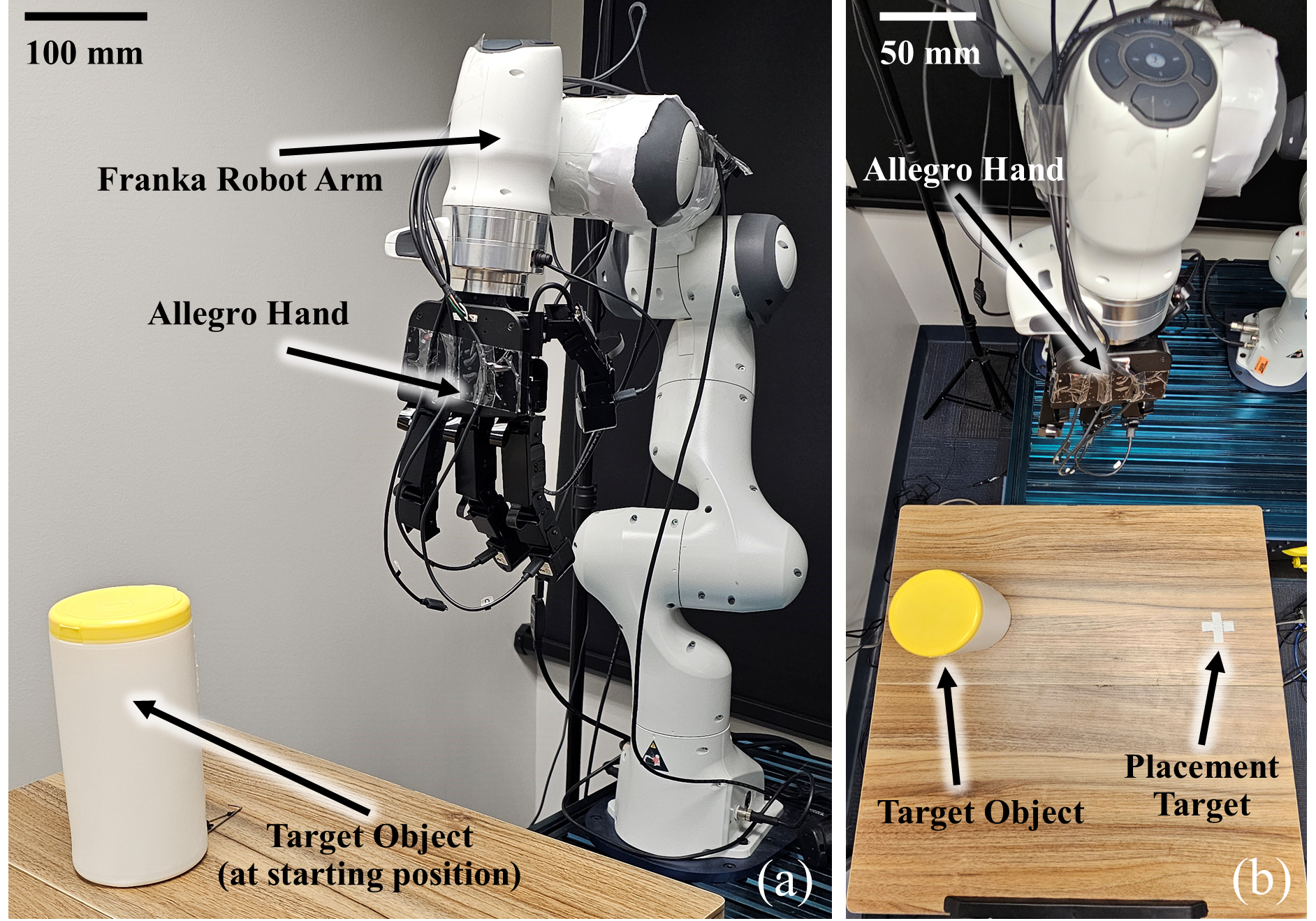}
    \caption{\textbf{Experiment setup Experimental Setup for the TEGA System Evaluation.} The setup consists of a 7-degree-of-freedom (7-DOF) Franka Emika Panda robotic arm with an Allegro Hand as the end-effector, equipped with DIGIT tactile sensors on each fingertip to capture real-time contact data. The human operator wears a motion tracker, three EMG sensors, and a bHaptics vest to facilitate teleoperation with haptic feedback. The operator’s forearm muscle contractions, detected by the EMG sensors, control the robotic hand’s grasping force, while the haptic vest provides vibrotactile feedback based on the Contact Concentration Index (CCI) and Effective Deformation Area (EDA) extracted from the tactile sensor data. The pick-and-place task is performed in a structured workspace.}
    \label{fig:5}
\end{figure}
\begin{figure}[!b]
    \centering
    \includegraphics[width=\columnwidth]{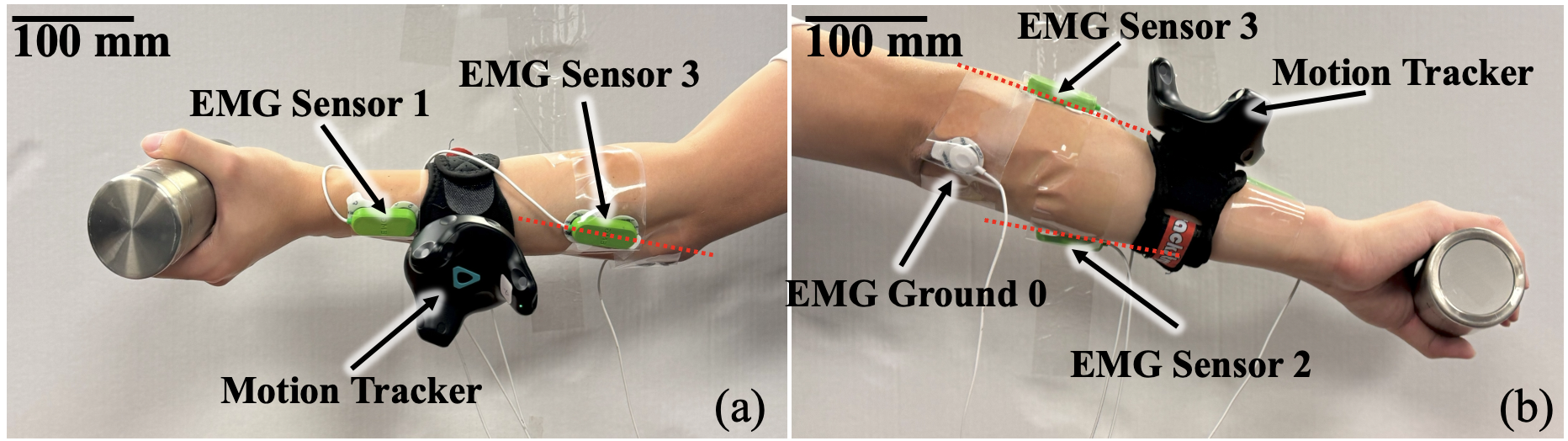}
    \caption{\textbf{EMG sensors and motion trackers mounted on the arm.} Placement of EMG Sensors Corresponding to Forearm Muscles. This figure illustrates the placement of three EMG sensors on the forearm, each corresponding to a specific muscle involved in grasp force modulation. Sensor 1 (S1) is placed over the Extensor Carpi Radialis (ECR), responsible for wrist extension and stabilization during grasping. Sensor 2 (S2) is placed over the Flexor Carpi Ulnaris (FCU), which assists in wrist flexion and contributes to grip force generation. Sensor 3 (S3) is placed over the Flexor Digitorum Superficialis (FDS), a key muscle responsible for flexing the fingers during grasp execution.}
    \label{fig:6}
\end{figure}

\begin{figure}[!b]
    \centering
    \includegraphics[width=\columnwidth]{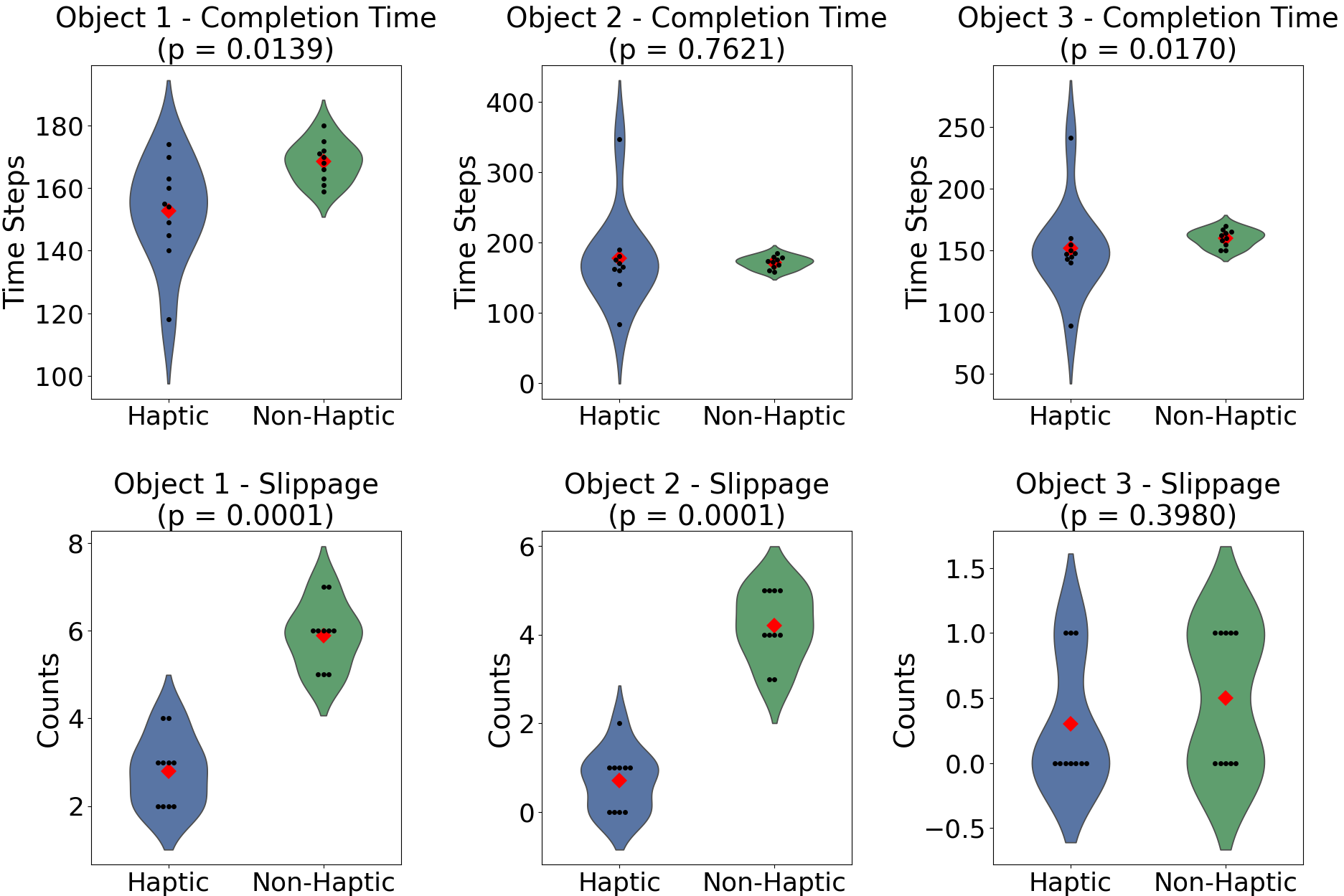}
    \caption{\textbf{Comparison of task performance metrics across Haptic and Non-Haptic condition.} Top row: completion time for Objects 1–3; bottom row: slippage counts. Plots show trial data (black dots), distributions (violin), and group means (red diamonds). Mann–Whitney U test p-values are indicated.}
    \label{fig:7}
\end{figure}

\begin{figure*}[!ht]
    \centering
    \includegraphics[width=\textwidth]{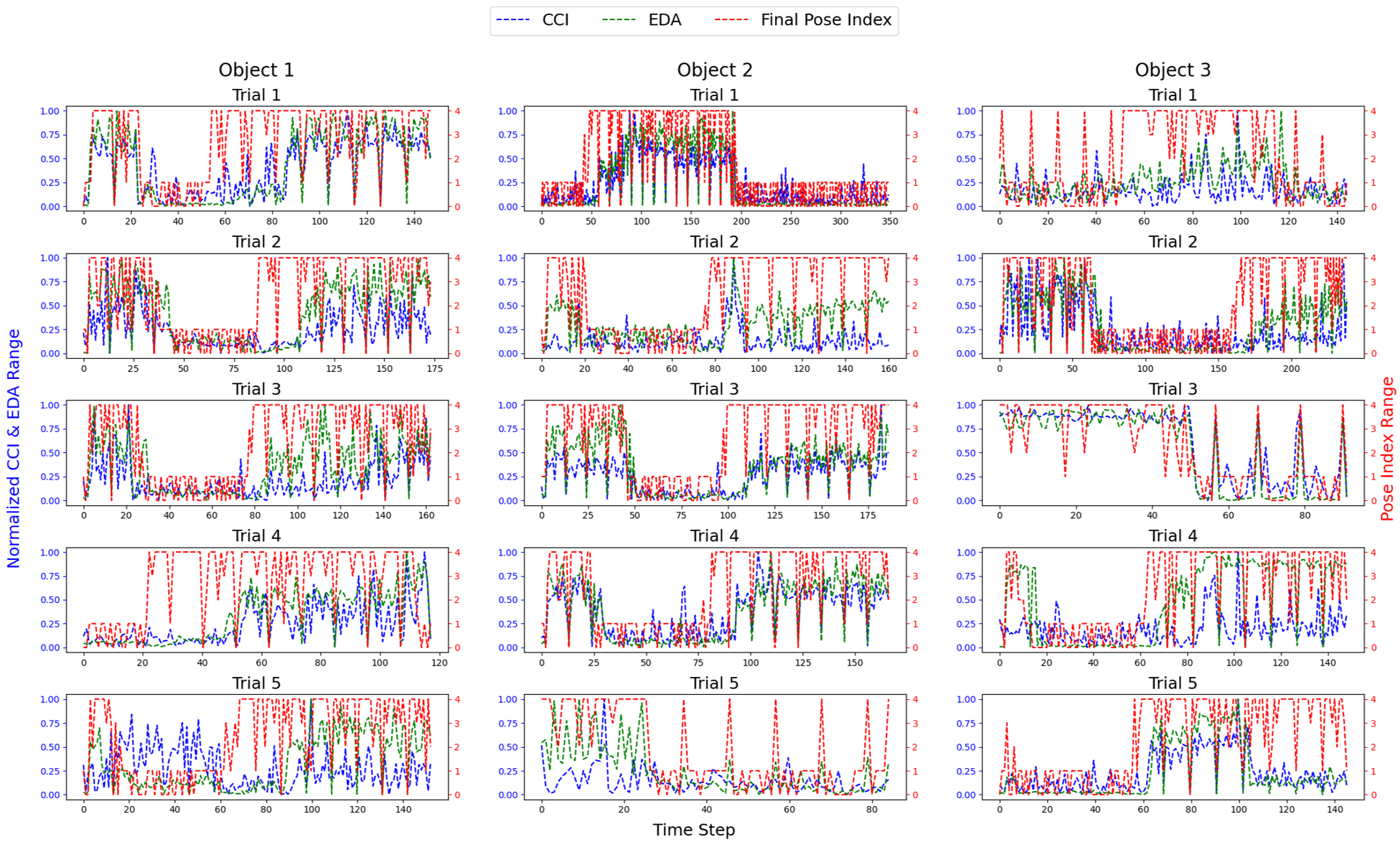} 
    \caption{
    \textbf{Real-Time Correlation Between Tactile Feedback and Grasping Pose Selection.} This figure illustrates the relationship between grasping pose selection and tactile feedback signals, including CCI and EDA, across different object types. The red dashed line represents the grasping pose index, indicating the level of applied force, where a higher index corresponds to a stronger grip. The blue solid line represents CCI, reflecting the sharpness of pressure concentration at the contact point. A higher CCI value indicates a more localized, intense contact. The green solid line represents EDA, which quantifies the spatial extent of the deformation. A larger EDA value suggests a broader contact area, while a smaller value indicates localized force application. Each column represents a specific object (Object 1, Object 2, Object 3), while each row corresponds to individual trials.
    }
    \label{fig:8}
\end{figure*}

Fig.~\ref{fig:5} shows the setup of the case study. The operator was required to control the position and orientation of the allegro hand mounted on a 7-degree-of-freedom (7-DOF) robotic arm (Franka Emika Panda) using forearm movements tracked via motion trackers. Robotic hand actuation was controlled using forearm muscle contractions detected by three EMG sensors placed on the user’s forearm. Note that although a motion tracker is mounted on the operator’s hand, the hand pose is not utilized and has no effect on the control logic, ensuring that the system accurately reflects the scenario of a limb-loss operator. Based on the haptic feedback, the operator adjusted their muscle contractions to maintain an optimal gripping force, either with or without haptic feedback. Fig.~\ref{fig:6} shows the placement of EMG sensors and motion trackers on the participant’s forearm.

The experiment followed a within-subjects design with ten repeating trials for each object. The two experimental conditions were: (1): EMG-VR control with haptic feedback; (2): EMG-VR control without haptic feedback. During each trial, the operator first controlled the robotic hand to approach the target object and then attempted to grasp it with minimal gripping force, applying only the necessary friction to lift the object without overexertion or deformation. To mitigate muscle fatigue, a one-minute rest period was provided between trials. Only the dominant forearm was used for the task.

\subsection{Performance and Analysis}
To evaluate the effectiveness of the proposed system under the experimental protocol described in Section IV-A, we assessed performance using three metrics: grip success rate, slippage count, and deformation. Grip success was defined at the task level if the object was lifted from the table and stably raised to a height of 0.2m within a fixed time window (15s) for a trial. Trials were labeled unsuccessful if the object was dropped or failed to reach the target height. Slippage events were identified through visual inspection by a single trained observer and recorded when clearly observable relative motion between the object and fingers occurred after lift-off. Deformation events were recorded when clearly visible over-compression or sustained shape distortion was observed during grasping. All slippage and deformation annotations were performed by the same observer to ensure consistency across trials.

\subsubsection{Overall Analysis}
Table~\ref{tab:1} and Fig.~\ref{fig:7} report the comparative performance of haptic versus non-haptic feedback. Completion times were largely comparable across conditions, with significant improvements observed for Object 1 (p = 0.014) and Object 3 (p = 0.017), but not for Object 2 (p = 0.763). Slippage counts showed the clearest benefit of haptic feedback: Object 1 and Object 2 exhibited substantial reductions (p $\leq$ 0.001 for both), confirming that haptic cues enabled better regulation of grip force to prevent unintended slipping. For Object 3, slippage remained minimal across both groups (p = 0.398). In addition, deformation counts for Object 3, reported in Table~\ref{tab:1}, were lower with haptic feedback, and the difference reached significance (p = 0.030), suggesting improved control when manipulating deformable objects.
\begin{table}[!b]
\centering
\caption{Performance across Haptic and Non-Haptic Conditions}
\label{tab:1}
\begin{tabular}{ccccc}
\toprule
\multirow{2}{*}{\textbf{Condition}}  & \multirow{2}{*}{\textbf{Metric}} & \textbf{Object 1} & \textbf{Object 2} &\textbf{Object 3} \\
 & & \textbf{Mean} & \textbf{Mean} & \textbf{Mean} \\
\midrule
\multirow{4}{*}{Haptic}     
& Completion Time & 152.8 & 177.4 & 151.5\\
& Slippage        & 2.8 & 0.7 & 0.3\\
& Deformation        & 0.0 & 0.0 & 0.5\\
& Success Rate & 5/5 & 5/5 & 5/5 \\
\midrule
\multirow{4}{*}{Non-Haptic} 
& Completion Time & 168.5 & 171.4 & 160.1\\
& Slippage        & 5.9 & 4.2 & 0.5\\
& Deformation      & 0.0 & 0.0 & 1.3\\
& Success Rate & 4/5 & 4/5 & 5/5 \\
\bottomrule
\end{tabular}
\end{table}

\subsubsection{Real-time Correlation }
To evaluate the effectiveness of the proposed tactile feedback module, we examined the relationship between grasping pose selection (red dashed line) and real-time tactile feedback from CCI (blue) and EDA (green). The goal was to assess whether the feedback mechanism dynamically reflected grasping force changes and provided actionable cues to the user. Fig.~\ref{fig:8} illustrates the temporal relationship between pose selection and tactile signals, where the x-axis denotes time steps, the left y-axis shows normalized CCI/EDA values, and the right y-axis represents the grasping pose index. Each column corresponds to a tested object, and each row corresponds to an individual trial.

It is important to note that the grasping command is discretized into five pose levels (0–4) rather than a binary on–off signal, enabling users to select intermediate force levels during manipulation. Beyond visual inspection, we further computed the Pearson correlation coefficient between the grasping pose index and normalized CCI/EDA signals. Across all trials, these correlations were consistently positive, indicating that increases in tactile feedback signals were associated with stronger grasping poses, while decreases corresponded to looser grips.

This synchronized behavior suggests that users actively responded to haptic cues by adjusting their grasping poses in real time. While the pose representation is discretized, the observed correlations and frequent transitions involving intermediate pose levels indicate graded and responsive force regulation, rather than simple on–off grasping. These results demonstrate the system’s ability to support adaptive grasp stabilization.

\section{DISCUSSION AND LIMITATIONS}

The results indicate that the proposed TEGA system supports more effective grasping force regulation by integrating EMG-based intent detection with real-time haptic feedback. While task completion times were largely comparable across conditions, haptic feedback significantly reduced slippage for rigid objects and excessive deformation for deformable objects, highlighting its benefit for grasp stability and safe force application.

Overall, haptic feedback improved success- and stability-related performance rather than execution speed, enabling users to regulate grasping force more reliably than EMG-based control alone. This suggests that the proposed dual-loop architecture can compensate for the lack of natural tactile sensation and support robust grasping across objects with varying properties.

Several limitations should be noted. The current system employs a rule-based EMG-to-force mapping, and cognitive load associated with vibrotactile feedback was not explicitly evaluated. The experiments were conducted with able-bodied participants using a limited set of objects and vertical grasp-and-lift motions, without independently varying object weight and stiffness or exploring diverse grasp strategies. In addition, slippage and deformation events were identified through visual observation, introducing a degree of subjectivity despite consistent annotation by a single observer. Moreover, EMG signals were used to infer user intent rather than to model the physical effects of object properties on muscle activation. Future work will explore learning-based control strategies, objective sensing-based evaluation, broader user studies, and more diverse grasping scenarios to improve personalization and generalizability.

\section{CONCLUSIONS}
This study demonstrates that integrating EMG-based intent detection with real-time haptic feedback enables adaptive grasping force control, improving grip precision and stability across objects with varying properties. Haptic feedback significantly reduced slippage for rigid objects and excessive deformation for soft objects, confirming its role in fine-tuning force modulation. While the system proved effective, further research is needed to assess user adaptability and cognitive load, as well as to refine the EMG-to-force mapping using learning-based models trained on larger datasets. Future work will involve expanded human trials to optimize system performance and enhance usability, ensuring more precise and intuitive control for individuals with upper limb disabilities or sensory impairments.

\addtolength{\textheight}{-12cm}   





\bibliographystyle{IEEEtran}

\bibliography{refs}

\end{document}